# Comparing Semantic Navigation in Humans and Large Language Models using Natural Language Processing

**Gabriel Paris-Colombo (gabrielparis461@gmail.com)[1], Rodrigo M. Cabral-Carvalho[1] & Felipe D. Toro-Hernández[1*]**

[1]Center for Mathematics, Computing and Cognition, Federal University of ABC

[1]{gabrielparis461,rodrigodamottacc,ftoroher}@gmail.com

***Corresponding author: ftoroher@gmail.com**

## Abstract

Semantic memory retrieval can be conceptualized as navigation through conceptual space. We compared semantic search dynamics between humans and three large language models (GPT-4o, Gemini-2.5-Pro, Claude-Sonnet-4.5) using verbal fluency data. By applying trajectory-based NLP metrics to the items generated by 82 human participants and LLM output across eight temperature settings, we quantified three complementary dimensions: entropy (step size predictability), distance to next (successive semantic steps), and distance to centroid (global dispersion). Humans exhibited higher entropy, larger semantic steps and broader dispersion than all LLMs, indicating more variable and exploratory search. Temperature tuning produced only partial alignments, as individual metrics matched between humans and LLMs at specific settings, but no configuration reproduced the complete human profile (in all dimensions). These findings suggest that human semantic search implements a distinctive balance between local exploitation and global exploration that current model architectures fail to reproduce.

## Introduction

Semantic memory is the long-term storage of general world knowledge, facts, concepts, and meanings (Tulving, 1972; Ralph et al., 2017; Binder & Desai, 2011). Its retrieval can be conceptualized as a navigational process with complex internal dynamics. To investigate the mechanisms underlying this search, researchers have commonly employed the Verbal Fluency Task (VFT), where participants generate as many items as possible from a category (e.g., “animals”) or beginning with a letter within a fixed interval, typically 60 seconds (Bousfield & Sedgewick, 1944). Fluency performance has often been summarized by the total number of words produced (Lezak et al., 2012; Shao et al., 2014; Strauss et al., 2006). However, semantic navigation accounts additionally leverage the temporal organization and semantic structure of responses to characterize how retrieval proceeds and to motivate theoretical models of semantic search (Troyer et al., 1997; Hills et al., 2012; Abbott et al., 2015).

A highly influential framework for understanding semantic fluency is the dual-component model (Troyer et al., 1997), which posits that performance depends on two dissociable processes: clustering (producing words within semantic subcategories, e.g., "pets") and switching (shifting between subcategories). Hills et al. (2012) extended this view by framing semantic retrieval as internal foraging, where effective search balances exploitation (local clustering) and exploration (switching to new semantic regions). Granular search models aim to specify the moment-to-moment mechanisms of retrieval (how agents transition through semantic space and when they make local versus long-range jumps). These include Random Walk models, which capture probabilistic transitions over semantic networks; Lévy Walk variants, which combine short local steps with occasional long jumps; and Area Restricted Search, which proposes that switching emerges from adaptive estimation of local "patch value" based on recent retrieval success (Morales et al., 2025; Rhodes & Turvey, 2007). While these models have been developed primarily to explain human semantic search, recent advances in artificial intelligence offer new opportunities to test whether similar dynamics emerge in systems that navigate semantic space through fundamentally different computational mechanisms.

Large Language Models (LLMs) are a class of artificial intelligence systems built on Transformer architecture (Vaswani et al., 2017), typically pretrained to predict the next token (word or subword unit) in sequences across large text corpora. This training objective leads them to learn statistical regularities that support fluent language generation. Internally, LLMs encode tokens as high-dimensional vectors and transform these representations across layers through self-attention mechanisms (Vaswani et al., 2017), which enable the extraction of multi-scale dependencies beyond mere lexical disambiguation. These architectures encode complex semantic hierarchies that encapsulate the structural and conceptual essence of entire phrases (Jawahar et al., 2019; Devlin et al., 2019), such that tokens with related meanings occupy nearby regions in the model's representational space (e.g., "dog" closer to "cat" than "airplane"). This property enables quantitative analyses of semantic structure and semantic movement during generation (Kalyan et al., 2021). Beyond their practical success across Natural Language Processing (NLP) tasks, LLMs have become increasingly relevant to cognitive science because they generate coherent, context-sensitive language that sometimes mirrors qualitative patterns of human linguistic behavior (Dasgupta et al., 2022; Manning et al., 2020). This has motivated proposals to treat LLMs as experimental comparators "silicon participants" whose outputs can be analyzed with the same behavioral metrics used for humans (Sarstedt et al., 2024). However, interpreting LLM behavior as cognitive evidence raises important methodological concerns, as these models often fail to replicate human strategic behavior in simple experiments and struggle to generalize across semantic variations that humans process intuitively (Gao et al., 2025; Schröder et al., 2025). Furthermore, LLMs' responses exhibit significant instability based on minor phrasing changes and rely on statistical token regularities rather than authentic psychological simulation (Gao et al., 2025; Schröder et al., 2025).

Recent work has examined whether LLMs can serve as useful behavioral comparators in cognitive tasks and to what extent their internal representations align with human semantic representations. For example, Hansen and Hebart (2022) showed that GPT-3 can generate semantic feature norms for a large set of objects that resemble human feature norms in the distribution of feature types (e.g., taxonomic, functional, encyclopedic) and that these features can predict human judgments of similarity, relatedness, and category membership. Furthermore, LLMs show sensitivity to human category structures, performing well in basic-level category induction, though they often struggle with more abstract superordinate levels (Pedrotti et al., 2025). Recent neuroscientific investigations using functional neuroimaging have provided additional evidence for this alignment: Ren et al. (2024) demonstrated that pre-training data size, model scaling, and alignment training positively correlate with LLM-brain similarity (i.e., the similarity between LLM representations and brain cognitive language processing signals), with instruction-tuned models showing significantly stronger alignment with fMRI signals during language processing. Lei et al. (2025) further showed that intermediate layers of LLMs align more strongly with brain activity than final layers, particularly in language-selective regions, and that this alignment respects hemispheric lateralization patterns observed in human language processing. Critically, when semantic search dynamics are compared using networks derived from fluency tasks, systematic structural differences emerge. Relative to humans, LLM-generated networks tend to show weaker local clustering and poorer global integration (Qiu et al., 2025), with semantic associates being more dispersed and organized into more isolated subgroups in semantic fluency networks (Wang et al., 2025). These patterns are consistent with less efficient navigation through semantic space, motivating trajectory-based step-by-step comparisons between humans and LLMs.

Despite recent advances in mapping the structural properties of LLM-derived semantic networks, a key gap remains in quantifying how these systems progress through semantic content on a step-by-step basis. Prior work has reported systematic differences between human and LLM semantic networks, such as lower clustering and higher modularity for the language models (Wang et al., 2025). Furthermore, while previous studies have successfully employed sequential distance metrics between static word embeddings to evaluate cognitive states in verbal fluency tasks (Linz et al., 2017; van Noort et al., 2025), these approaches assess semantic transitions by treating each generated word independently of its broader retrieval history. The present work extends the computational framework introduced by Toro-Hernández et al. (2026) by using their physics-inspired NLP metrics to assess semantic representations in LLMs. Instead of relying on sequential distances between static word representations, this approach utilizes sequential distances between cumulative contextual embeddings. We specifically chose this cumulative approach because it allows the semantic state to unfold over time, better approximating the accumulating working memory demands and contextual constraints inherent to human memory search (Baddeley, 2000). Following Toro-Hernández, Vieira & Cabral-Carvalho (2026), we quantify (i) distance to next, capturing semantic "jumps" between successive properties; (ii) semantic entropy, summarizing the predictability of step sizes across the sequence; and (iii) distance to centroid, indexing the degree of clusterization. Together, these metrics provide a fine-grained characterization of semantic navigation that complements former approaches.

With these analyses, we aim to characterize systematic differences in semantic navigation between humans and LLMs. Guided by prior work, we first expect LLM outputs to vary across temperature settings. Because temperature modulates sampling stochasticity during generation (Li et al., 2025), we consider that higher temperatures will yield more variable trajectories, reflected in larger distance to next values (broader "semantic jumps" between consecutive items), higher semantic entropy (greater diversity in exploration), and increased distance to centroid (production of more peripheral category exemplars), whereas lower temperatures will produce more predictable, potentially more rigid trajectories with reduced semantic variation. We further expect the closest correspondence to human navigation to emerge at intermediate temperatures, where generation is neither overly deterministic nor excessively noisy. Across model families, we anticipate that LLMs will generally show higher distance to next and entropy values than humans, consistent with their more dispersed semantic organization (Qiu et al., 2025; Wang et al., 2025), though meaningful variation may emerge due to differences in training and alignment procedures (Ren et al., 2024; Lei et al., 2025). Finally, to benchmark the generality of these effects across model families, we repeat the fluency-generation procedure with three state-of-the-art LLMs: gpt-4o, gemini-2.5-pro, and claude-sonnet-4.5, and compare the resulting trajectory profiles. By comparing temperature settings and model configurations, we test which conditions yield trajectory metrics most similar to humans, and we use these matches to constrain candidate mechanisms underlying human-like semantic navigation. This approach provides a framework for generating testable hypotheses regarding the constraints and search dynamics of human semantic memory.

# Methods

## Participants

To compare LLM-generated fluency with human performance, we used behavioral data from the SNAFU database (Zemla et al., 2020). In this dataset, 82 participants completed a classic semantic verbal fluency task across multiple categories. For each category, participants were instructed to produce as many items as possible within a 3-minute interval, avoiding repetitions and avoiding responses that shared the same suffix. In the present analyses, we focused on the animals category.

## LLM-generated fluency

We evaluated three LLMs (gemini-2.5-pro, gpt-4o, and claude-sonnet-4.5) on a semantic fluency task. Following Wang et al. (2025), we used a role-playing induction to increase output diversity and reduce repetitive list-like completions (Shanahan et al., 2023; Zhao et al., 2024). We used the 30 different occupations described by Wang et al. (2025). For each trial, the model was first prompted to describe the characteristics of a target occupation ("Do you know a/an [occupation]? Tell me the characteristics of a/an [occupation].") and then instructed to adopt the role ("Now you need to pretend that you are a person with the above characteristics completely. All the following answers must be in accordance with the above characteristics."). After this induction, the model was asked to generate an animal-fluency list (one animal per line), using standard verbal-fluency constraints (Henry & Crawford, 2004).

To reduce incomparability introduced by human timing constraints versus model generation time (Ding & Wang, 2025; Jain et al., 2023), we did not impose a time limit on the models. Instead, for each LLM trial, we defined a target list length by sampling from the empirical distribution of the number of items produced by human participants in the SNAFU animals category (Zemla et al., 2020). This matched-length stopping rule yields model-generated sequences that are comparable to human sequences in overall output size while avoiding assumptions about the equivalence of response latencies across humans and artificial agents. Model outputs were formatted as one item per line into an Excel sheet to facilitate parsing and downstream analyses.

To examine how stochasticity during generation affects semantic navigation, we varied temperature across trials (Herrera-Poyatos et al., 2025). Temperature modulates sampling randomness, with lower values typically producing more deterministic outputs and higher values producing more diverse outputs (Li et al., 2025; Renze, 2024). Accordingly, model temperature was varied to capture a broader range of responses (Qiu et al., 2025). Each LLM generated responses at each of eight temperature settings: 0.0, 0.15, 0.3, 0.45, 0.6, 0.75, 0.9, and 1.0, using each of the occupations.

## Natural language processing analyses

To analyze semantic representations in both human and LLM-generated responses, we used NLP methods to derive metrics that capture the step-by-step granularity of semantic navigation (Nour et al., 2023; Toro-Hernández et al., 2024). We use transformer-based text embeddings, which map text to high-dimensional vector representations that capture contextual meaning (Vaswani et al., 2017). Following recent practices in computational semantics (Hansen & Hebart, 2022; Wang et al., 2025; García et al., 2025; Sanz et al., 2022), we used OpenAI's text-embedding-3-large to encode each produced item into a shared latent semantic space. Treating responses as points in this space enables rigorous quantification of relationships between successive outputs; in turn, distance-based measures can be used to characterize navigational patterns and the semantic structure of the trajectories generated by humans and models (Toro-Hernández et al., 2026). To ensure robustness and address potential concerns about reliance on proprietary embedding models, all metrics were validated using multiple embedding systems, including both commercial and open-source alternatives.

For each verbal fluency trial, we utilized these high-dimensional representations to calculate three complementary metrics (Toro-Hernández et al., 2026). These measures quantify the proximity between successive responses and the overall density of the evoked features, providing a multidimensional view of memory search performance. Importantly, verbal fluency is sequential and depends on executive processes such as maintaining task goals, monitoring prior responses to avoid repetition, and inhibiting intrusions, functions linked to working-memory maintenance and cognitive control (Baddeley, 2000; Troyer et al., 1997). To approximate this accumulating contextual demand, trajectory-based metrics were implemented with a cumulative embedding representation, by which, at step t, the semantic state summarizes the items generated up to that point by combining the current item embedding with embeddings of all previously generated items (Toro-Hernández et al., 2026). This cumulative formulation allows semantic navigation dynamics to be modeled as they unfold over time, rather than treating each response as independent. In contrast, the distance to centroid metric was computed from non-cumulative item embeddings, because it is intended to index dispersion (semantic density) relative to an individual's overall semantic context rather than the evolving cumulative state (Toro-Hernández et al., 2026).

## Natural language processing metrics

To assess step-by-step semantic navigation dynamics, we extracted three NLP metrics from Toro-Hernández et al. (2026) (see that paper for full mathematical details).

**Distance to next.** We mapped each response stream to a semantic trajectory $X = (x_1, \ldots, x_N)$ using cumulative embeddings, where $x_t$ encodes the concatenated prefix of items 1:t (e.g., if the first two items are "cat" then "dog," $x_2$ embeds "cat dog"). We then computed distance to next as the cosine distance between successive unit-normalized states, $x_t$ and $x_{t+1}$, yielding an N - 1 series of step sizes ("semantic jumps"). Larger values indicate larger moment-to-moment shifts in semantic state (e.g., "dog" → "shark" > "cat" → "dog"). For a length-invariant summary, we used the mean step size across all steps in the trajectory.

**Entropy.** To quantify the predictability of step-size dynamics, we computed a length-normalized Shannon entropy over the distance to next series (Shannon, 1948). Within each trajectory, step sizes were binarized as "high" vs "low" relative to the within-trajectory median, producing a binary sequence from which Shannon entropy was computed and normalized for the number of valid steps. Entropy was only computed for sequences with at least three steps. Higher entropy indicates a less predictable, more variable pattern of step sizes across the trajectory.

**Distance to centroid.** To index overall dispersion (clustering vs spread) relative to an individual's central semantic context, we computed a centroid-based measure from non-cumulative item embeddings. Repeated items were collapsed so redundancy did not overweight specific responses; for each unique item, we retained the embedding from its first occurrence and defined the centroid as the average of all unique-item embeddings. Each item was then assigned the cosine distance between its embedding and this centroid. Higher distance to centroid values indicate a more dispersed semantic set, whereas lower values indicate a more focused (more clustered) set of responses.

## Data analysis

We compared human participants with LLM-generated responses across eight temperature conditions (0.0, 0.15, 0.30, 0.45, 0.60, 0.75, 0.90, and 1.0) using generalized linear mixed-effects models (GLMMs). For each NLP metric, we fitted a model with group as a fixed effect and participants/instance ID as a random effect to account for repeated measures and individual variability. Models were fitted according to the most appropriate distribution, assessed using DHARMa residual diagnostic in R. Analyses were performed with the glmmTMB package (Brooks et al., 2017) in R (v.4.3.1).

# Results

Our analyses proceeded in two stages. First, we established a baseline by comparing general performance metrics between each group (gpt-4o, gemini-2.5-pro, claude-sonnet-4-5, human). Second, we conducted a series of contrasts between the human group and LLMs groups performance at each specific temperature level. Full statistical reports for all comparisons, including estimate values and exact p-values, are provided in Supplementary Materials (available at https://osf.io/gkp57)

## Category comparison

**Entropy** Human participants exhibited substantially higher entropy than all three LLMs. Among LLMs, Claude-sonnet-4.5 showed lower entropy than both Gemini-2.5-pro and GPT-4o, whereas Gemini and GPT-4o did not differ reliably from one another (see Table 1 for comparisons).

**Distance to next** Humans produced larger semantic steps than each LLM. Among LLMs, Claude-sonnet-4.5 showed the smallest step sizes, differing reliably from Gemini-2.5-pro but not from GPT-4o.

**Distance to centroid** Human responses were the most semantically dispersed. All three LLMs showed more centroid-focused responses, with GPT-4o exhibiting the strongest clustering, followed by Gemini-2.5-pro. All pairwise contrasts were significant.

**Summary of category comparisons** Human semantic trajectories exhibit greater variability and less predictable step-size dynamics, whereas all LLMs show more regular, constrained entropy profiles consistent with a more rigid navigation regime. Similarly, human navigation involves larger successive semantic shifts between items, whereas LLMs exhibit more “sticky” local trajectories with tighter moment-to-moment transitions through semantic space, particularly for Claude. At a broader scale, LLM responses maintain a more conservative global search profile, exploring a narrower semantic breadth compared to the wider dispersion characteristic of human semantic retrieval.

Table 1: Tukey results for category comparisons.

| Metric | Contrast | Δ | p |
|---|---|---|---|
| Entropy | Claude vs. Human | -0.211 | <0.001 |
| | Gemini vs. Human | -0.187 | <0.001 |
| | GPT vs. Human | -0.192 | <0.001 |
| | Claude vs. Gemini | -0.024 | <0.001 |
| | Claude vs. GPT | -0.019 | 0.003 |
| | Gemini vs. GPT | 0.005 | 0.732 |
| Distance N | Claude vs. Human | -0.106 | <0.001 |
| | Gemini vs. Human | -0.068 | 0.003 |
| | GPT vs. Human | -0.085 | <0.001 |
| | Claude vs. Gemini | -0.038 | <0.001 |
| | Claude vs. GPT | -0.021 | 0.096 |
| | Gemini vs. GPT | 0.017 | 0.257 |
| Distance C | Claude vs. Human | -0.025 | <0.001 |
| | Gemini vs. Human | -0.016 | <0.001 |
| | GPT vs. Human | -0.058 | <0.001 |
| | Claude vs. Gemini | -0.010 | <0.001 |
| | Claude vs. GPT | 0.033 | <0.001 |
| | Gemini vs. GPT | 0.043 | <0.001 |

## Human vs. LLMs temperatures

**Entropy** Results showed that alignment with humans was model-specific (Figure 1 - A): Claude-sonnet-4.5 aligned most closely at high-intermediate temperatures (e.g. T = 0.75 and T = 0.90), whereas Gemini-2.5-pro and GPT-4o aligned most clearly at T = 0, with additional convergence at T = 1.0 and T = 0.60.

**Distance to next** Here, results were also model-specific (Figure 1 - B). Claude-sonnet-4.5 converged toward the human baseline at higher intermediate temperatures (T = 0.75 and 0.90), showing reliably smaller step sizes at lower temperatures. Gemini-2.5-pro showed the most human-like settings at the endpoints and some lower temperatures (T = 0, 0.15, 0.60, 0.75, 1.0), with reduced "semantic jumping" at T = 0.30, 0.45, and 0.90. GPT-4o was indistinguishable from humans at T = 0, 0.45, 0.60, and 1.0, but showed more locally constrained transitions at T = 0.15, 0.30, 0.75, and especially 0.90.

**Distance to centroid** Global dispersion was the hardest dimension to match (Figure 1 - C). Only a narrow setting for Claude (T = 0.60) and a mid-range band for Gemini (T ≈ 0.30 to 0.90, excluding endpoints) were statistically indistinguishable from humans. GPT-4o remained persistently non-human-like across all temperatures, showing consistently reduced distance to centroid values.

**Summary of human vs. temperature comparisons** Overall, the temperature comparisons show that “human-like” behavior does not emerge as a simple monotonic function of added sampling noise; instead, it depends on how each architecture converts temperature into semantic variation. This is most apparent for local trajectory dynamics such as entropy and distance to next, where models exhibit distinct alignment regimes, with Claude approaching humans at higher intermediate temperatures, while Gemini and GPT-4o show more irregular convergence patterns with pockets of locally constrained transitions. By contrast, global dispersion (distance to centroid) is harder to recover through temperature tuning, suggesting that global semantic breadth is limited by deeper representational constraints that temperature alone may not overcome.

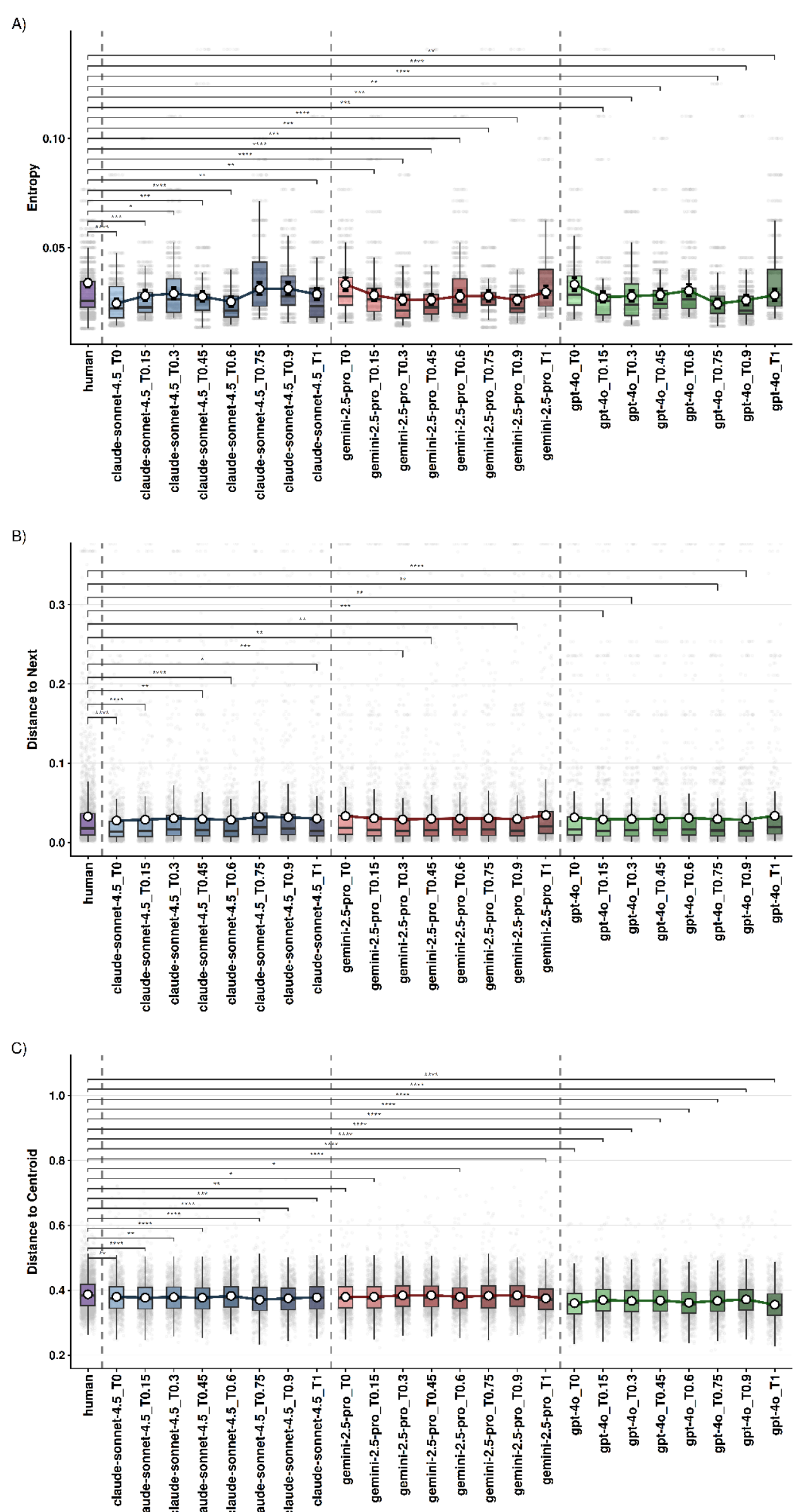

**Figure 1. Human baseline vs. all LLM–temperatures.** Entropy (A), Distance N (B), and Distance C (C). The leftmost group shows humans, followed by LLM outputs grouped by model; Claude Sonnet 4.5; Gemini 2.5 Pro; and GPT-4o. Vertical dashed lines separate model families. Gray points show individual observations; boxplots show medians and interquartile ranges. Circles indicate estimated marginal means (± 95% CI); within each model family, line color intensity increases with temperature (lighter = T = 0.0, darker = T = 1.0). Brackets denote Tukey-adjusted contrasts against the human baseline (* $p < .05$, ** $p < .01$, *** $p < .001$, **** $p < .0001$). All plots in high quality are provided in Supplementary Materials.

## Discussion

Our findings show reliable differences in how humans and LLMs navigate semantic space in verbal fluency. Contrary to expectations derived from prior reports of broader dispersion in LLM semantic neighborhoods (Qiu et al., 2025; Wang et al., 2025), our trajectory-based metrics reveal the opposite pattern. Humans exhibited higher entropy (less predictable step-size dynamics), larger successive semantic shifts, and broader dispersion around the centroid, whereas all three LLMs showed more regular, locally clustered, and centroid-focused trajectories. Consistent with optimal foraging accounts, human search appears to balance exploitation of local patches with exploration of distant regions (Hills et al., 2012), while LLM outputs appear more biased toward high-probability continuations learned from training data (Lake & Murphy, 2023; Bender et al., 2021). Accordingly, human-like semantic search is not recovered by tuning a single parameter; it requires joint alignment across unpredictability, step magnitude, and global dispersion.

As expected, LLM outputs varied across temperature conditions. However, alignment with humans was architecture-specific. Gemini consistently occupied an intermediate position (less variable than humans but more exploratory than Claude or GPT-4o), GPT-4o showed the most extreme clustering (most conservative global search), and Claude showed the tightest local clustering. Even when collapsing across temperatures, all LLMs differed from humans, suggesting that architectural and training differences create systematic biases in semantic organization that persist regardless of sampling strategy. Through the lens of the dual-component model (Troyer et al., 1997), this pattern is consistent with reduced switching and over-reliance on clustering. Crucially, temperature tuning produced only fragmented alignment: individual metrics could be matched at particular settings, but no configuration reproduced the integrated human profile, with the strongest dissociation in global dispersion (distance to centroid), where GPT-4o remained constrained regardless of stochastic input.

These results challenge the notion that LLMs can serve as straightforward "silicon participants" (Sarstedt et al., 2024) in cognitive research. Most contemporary LLMs undergo extensive post-training with reinforcement learning from human feedback (RLHF) to optimize performance as assistants, not to faithfully replicate human cognitive processes (Ouyang et al., 2022; Bai et al., 2022), which may contribute to introducing systematic deviations from human behavior. In our data, temperature can make models appear human-like on isolated metrics, but not on the joint profile across entropy, step size, and dispersion, creating a central challenge for using LLMs as cognitive models, in which calibrating temperature to match one semantic dimension can distort others. This trade-off suggests that current instruction-tuned LLM architectures may make it difficult to simultaneously capture all dimensions of human semantic search. This also limits cross-study generalizability, because different metrics align with humans at different temperatures, so selecting temperature based on a single outcome may introduce systematic biases. Future work should validate parameter choices against multiple theoretically motivated dimensions rather than optimizing for one measure.

## The Challenge of LLMs as Psychological Surrogates

It is important to acknowledge methodological limitations when interpreting LLM behavior as psychological evidence. Recent studies show that LLMs often fail to reproduce human strategic behavior in simple experiments and struggle to generalize across semantic variations that humans handle intuitively (Gao et al., 2025; Schröder et al., 2025). Their responses can also be unstable under minor prompt changes and are driven by statistical regularities in token sequences rather than genuine psychological simulation (Gao et al., 2025; Schröder et al., 2025). Our findings contribute to this literature by showing that even when LLMs may produce superficially fluent outputs in semantic fluency, the underlying navigational dynamics diverge systematically from human retrieval. Relatedly, Lu et al. (2025) report that LLM-generated tasks are markedly less social and less physical than human-generated tasks and show a thematic bias toward abstraction.

From a theoretical perspective, LLMs are trained to capture statistical structure in text (Schüle, 2024), and they lack the embodied experience and goal constraints that shape human cognition (Gao et al., 2025). Schröder et al. (2025) further show that even CENTAUR (a model fine-tuned on millions of human responses across many psychological experiments; Binz et al., 2025) fails to preserve human behavioral patterns when items are semantically reworded, despite surface-level similarity. This parallels our observation that temperature manipulation, although it modulates output variability, does not fundamentally change the underlying navigational profile in a way that simultaneously matches humans across multiple dimensions of semantic search. Accordingly, researchers should be cautious when using LLM behavior as evidence for or against cognitive theories without extensive validation across tasks and parameter settings (Niu et al., 2024). Our results underscore this point: similarity on a single metric does not imply equivalence of the full behavioral signature, a dissociation that would be theoretically informative in human data but may instead reflect architectural constraints in current LLMs.

## Human Semantic Search: Unpredictability and Exploration

Our results also offer insights into human semantic memory search. The joint pattern of high entropy, large distance to next, and broad distance to centroid suggests that effective human foraging relies on a specific configuration of variability, one that supports exploration without devolving into random wandering. The entropy findings are particularly informative. Traditional accounts emphasize executive control mechanisms that monitor retrieval success and initiate switches when local patches are exhausted (Troyer et al., 1997; Rosen & Engle, 1997). Our data are consistent with the possibility that semantic navigation introduces controlled stochasticity into transitions, enabling occasional “surprising” moves that depart from purely associative retrieval. Elevated distance to next values reinforces this interpretation. Whereas LLMs tend to take smaller, locally clustered steps consistent with their training statistics, humans produce larger step sizes, aligning with Lévy-walk accounts of semantic search in which optimal foraging combines frequent short movements with occasional long jumps (Rhodes & Turvey, 2007). In this sense, human trajectories appear to implement a balance between local exploitation and unpredictability. Furthermore, the broader distance to centroid may indicate that human search covers diverse regions of semantic space in an efficient way, rather than making directionless leaps, consistent with executive control accounts in which working memory helps track retrieved items and supports strategic shifts across subregions of a category (Baddeley, 2000).

Several limitations warrant consideration. First, we examined only one semantic category. Since categories vary in internal structure, future research should explore potentially specific differences across categories in the human-LLM comparison. Second, while we used OpenAI's text-embedding-3-large and validated results with alternative systems, all embedding models are trained artifacts with particular biases. The human-LLM differences we observed could partially reflect misalignment between embedding space geometry and human semantic structure. Future work could explore embeddings derived from human behavioral data (e.g., semantic fluency norms, feature generation tasks, similarity judgments) that may better capture human semantic organization. Finally, our reliance on proprietary, closed-weight LLMs restricts mechanistic interpretation. Future studies should utilize open-weight, locally runnable models to directly investigate how internal layer representations correlate with behavioral outputs, offering deeper insights into LLM semantic navigation across a broader range of datasets and categories.

## Conclusion

This study offers important insights regarding differences in how humans and LLMs navigate semantic space during verbal fluency tasks, highlighting different searching patterns. While specific temperature settings can approximate human-like performance on isolated metrics, no configuration simultaneously captured the integrated signature of human semantic search. These findings challenge the use of current LLMs as straightforward cognitive models, suggesting that their optimization for assistant-like performance through RLHF may introduce systematic deviations from human cognitive processes. Future research should validate LLM parameters across multiple theoretically motivated dimensions and explore whether alternative training objectives or architectural modifications could better capture the computational principles underlying human semantic memory search.

## Acknowledgments

The authors used large language models to assist with limited aspects of code and manuscript editing. All AI-assisted material was checked and revised by the authors as needed. The authors take full responsibility for the final content of this article.

**Supplementary Materials** Supplementary materials for this article, including full statistical reports for all comparisons, are available at https://osf.io/gkp5